\documentclass[11pt,english]{article}
\usepackage{mathptmx}

\usepackage[T1]{fontenc}
\usepackage[latin9]{inputenc}
\usepackage{geometry}
\usepackage{float}
\usepackage{textcomp}
\usepackage{amsthm}
\usepackage{amsmath}
\usepackage{amssymb}
\usepackage{graphicx}

\makeatletter
\numberwithin{equation}{section}
\numberwithin{figure}{section}
  \theoremstyle{remark}
  \newtheorem*{acknowledgement*}{\protect\acknowledgementname}

\makeatother

\usepackage{babel}
  \providecommand{\acknowledgementname}{Acknowledgement}

\begin{document}

\title{Sequential testing over multiple stages and\\
performance analysis of data fusion}

\author{Gaurav Thakur%
\thanks{\noindent MITRE Corporation, McLean, VA 22102, email: gthakur@alumni.princeton.edu\protect \\
Approved for Public Release; Distribution Unlimited. 13-0855\protect \\
©2013-The MITRE Corporation. All rights reserved.%
}}

\date{February 18, 2013}
\maketitle
\begin{abstract}
We describe a methodology for modeling the performance of decision-level
data fusion between different sensor configurations, implemented as
part of the JIEDDO Analytic Decision Engine (JADE). We first discuss
a Bayesian network formulation of classical probabilistic data fusion,
which allows elementary fusion structures to be stacked and analyzed
efficiently. We then present an extension of the Wald sequential test
for combining the outputs of the Bayesian network over time. We discuss
an algorithm to compute its performance statistics and illustrate
the approach on some examples. This variant of the sequential test
involves multiple, distinct stages, where the evidence accumulated
from each stage is carried over into the next one, and is motivated
by a need to keep certain sensors in the network inactive unless triggered
by other sensors.\end{abstract}
\begin{acknowledgement*}
The author would like to thank Mr. Andrew Knaggs of the Joint IED
Defeat Organization (JIEDDO) for supporting and funding this work,
Dr. Tom Stark of JIEDDO for technical and operational guidance, and
Dr. Dave Colella and Dr. Garry Jacyna of MITRE for providing valuable
feedback and suggestions on the paper.
\end{acknowledgement*}

\section{Introduction\label{SecIntro}}

The JIEDDO Analytic Decision Engine (JADE) is a flexible software
toolkit for studying the performance of sensor configurations for
the detection of person-borne explosive compounds and other threat
substances. JADE is designed to enable performance and tradeoff analyses
between different, user-specified scenarios with given sensor placements
and data fusion networks. JADE contains fundamental physics-based
models of several sensor technologies of interest, such as nonlinear
acoustic and radar-based detectors, along with a data fusion system
that we focus on in this paper. The fusion system consists of a static
component that combines the decisions of individual sensors at a fixed
point in time, and a dynamic, time-dependent component that in turn
fuses the outputs of the static structure at different times. The
static component is based on a probabilistic graphical model, or Bayesian
network, and accepts probability matrices from the physics-based sensor
models as inputs (the details of which are abstracted from the fusion
system). Its outputs are fed into the dynamic fusion framework, which
is based on sequential hypothesis testing and produces performance
metrics for the entire, fused sensor configuration. The purpose of
the system is to determine the performance of a given fusion structure,
as opposed to doing fusion on actual measurements.\\

We first discuss the static framework in Section \ref{SecStatic},
which allows elementary fusion structures to be stacked and analyzed
efficiently. This material is fairly standard but serves as a background
for the rest of the paper. We then describe an extension of the Wald
sequential test in Section \ref{SecDynamic} that involves multiple,
distinct stages, where the evidence accumulated from each stage is
carried over into the next one. We show how the performance characteristics
and decision times of such a test can be computed efficiently for
time-dependent statistics and illustrate this approach on examples
in Section \ref{SecExamples}. This setup models a bank of anomaly
sensors that observe a moving target over time, reach an initial fused
decision, and if justified, activate additional sensors that continue
to collect static fused evidence over time until a final decision
is made about the target. The multiple-stage configuration allows
sensors that have a high cost of operation to remain inactive unless
specifically called upon.

\section{Static fusion using Bayesian networks\label{SecStatic}}

The static fusion structure is formulated as a Bayesian network, i.e.
a directed acyclic graph with each vertex representing a random variable
and edges describing dependencies between the variables. A Bayesian
network has the defining property that every vertex is conditionally
independent of its ancestor vertices given its immediate parent vertices
\cite{Pe97}. Such networks are an intuitive framework for performing
probabilistic inference among interconnected events in many different
contexts, and are well suited for formulating a sensor fusion system.
The vertices in our network represent the object, sensors and fusion
centers.\\

\begin{figure}[h]
\centering{}\includegraphics[scale=0.9]{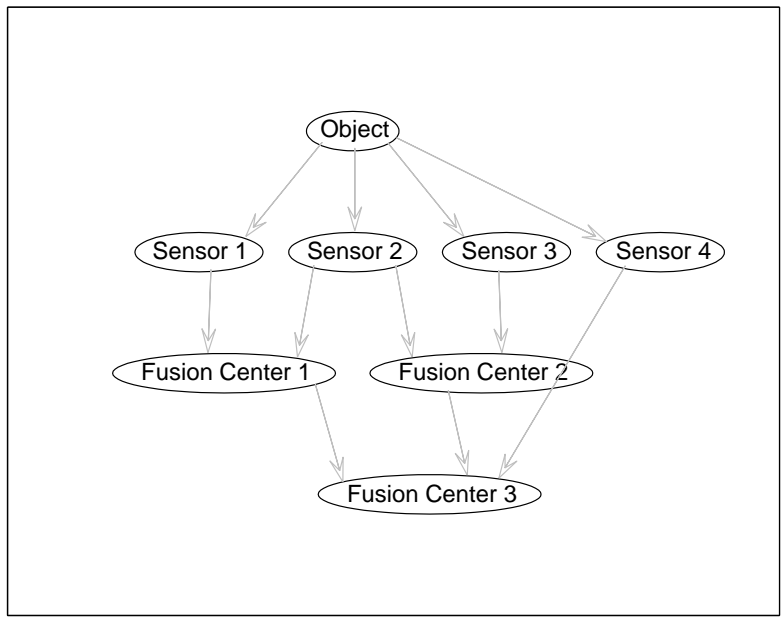}\caption{\label{FusionGraph}An example static fusion network.}
\end{figure}

Suppose we have $N$ sensors to be fused, each of which outputs hard
decisions between $M$ possibilities (with the first one corresponding
to the case where no threat is present). Let $H$ be the true object
(or the hypothesis in a Bayesian setting) and $S$ be the local decision
of a given sensor. The performance of the sensor is described by the
$M\times M$ matrix $\{P(S=m|H=m')\}_{1\leq m,m'\leq M}$, which we
write concisely as $P(S=\cdot|H)$. In this paper, we will generally
focus on $M=2$, corresponding to binary decision-level fusion, but
the discussion in this section applies to other $M$ as well. At any
fusion center $F$ with $V$ parent vertices $\{S_{n}\}_{1\leq n\leq V}$,
we can describe the fusion rule by the $V$-dimensional tensor $P(F=\cdot|\{S_{n}\}_{1\leq n\leq V})$,
which for deterministic fusion rules consists only of $0$ and $1$
elements. The performance of the entire system is given by $P(D=\cdot|H)$,
where $H$ is the root vertex in the graph and the system's final
decision $D$ is the last child vertex. This formulation enables the
graph to take on essentially any desired form and allows different
combinations of fusion centers and sensors to be stacked together,
subject to the following rules that ensure that the fusion structure
is meaningful.
\begin{itemize}
\item Each sensor vertex must have the object and at most one fusion center
as its parent.
\item At least one sensor must have only the object as its parent.
\item Each fusion center can have any combination of sensors and/or fusion
centers as parents, as long as no cycles are formed in the graph.
\item No fusion center can have the object as a parent.
\item There must be exactly one fusion center with no children, representing
the final decision.
\end{itemize}
These rules ensure that all sensors in the graph observe the object
and that all intermediate decisions are ultimately combined at a single,
final fusion center. An example fusion network of this type is shown
in Figure \ref{FusionGraph}.\\

To determine $P(D=\cdot|H)$, we choose small subgraphs of the Bayes
network at a time, each containing one fusion center and all its parent
vertices, and marginalize over them using the standard method of belief
propagation \cite{Pe85,Pe97}. This is done iteratively for each fusion
center in parent-to-child order until all the fusion centers have
been covered. For certain fusion rules, the probability matrix at
any child fusion centers may depend on the outputs of any parent fusion
centers, so this iterative procedure is much more simple and efficient
than having the child fusion centers' conditional probabilities account
for this dependence and computing marginal probabilities over the
entire Bayes network at once.\\

At each fusion center, JADE allows the user to choose between five
elementary hard-decision fusion rules: the ``and,'' ``or,'' majority,
Neyman-Pearson optimal and Bayes optimal rules \cite{Va96}. Any of
the five fusion rules can be used in the decision-level case of $M=2$,
while for $M>2$, only the Bayes and majority rules are meaningful.
At any given fusion center, let $\{S_{n}\}_{1\leq n\leq V}$ be the
local decisions of $V$ sensors feeding into it, with $0\leq S_{n}\leq M-1$,
and let $F$ be the fused decision. The ``and'' rule simply chooses
$F=1$ if all the $S_{n}=1$, and $F=0$ otherwise. Similarly, the
``or'' rule chooses $F=0$ if all the $S_{n}=0$, and $F=1$ otherwise.
It is clear that the ``and'' rule minimizes $P_{F}$ while keeping
$P_{D}>0$ and the ``or'' rule maximizes $P_{D}$ while keeping
$P_{F}<1$, so they can be thought of respectively as the least and
most sensitive fusion rules available. The majority rule takes a majority
vote between the sensors, i.e. $F=\mathrm{mode}(\{S_{n}\})$, with
a random, uniformly distributed decision taken if there is a tie between
multiple choices. This is the only rule where the fused decision is
potentially random. These three rules generally do not satisfy any
good optimality criteria, but are conceptually simple and useful as
a baseline for comparison against the two optimal rules.\\

At a given fusion vertex in the network, the Neyman-Pearson rule (for
$M=2)$ has the user specify a target false alarm probability $P_{F}'$,
and the system chooses the (deterministic) fusion rule that maximizes
the local $P_{D}$ at that vertex, subject to the constraint $P_{F}\leq P_{F}'$.
The optimal rule is found by computing the likelihood ratios $\prod_{n=1}^{V}\frac{P(S_{n}|H=1)}{P(S_{n}|H=0)}$
for every combination of individual sensor decisions $\{S_{n}\}$
and arranging them in increasing order. The combinations are then
partitioned into two subsets $I$ and $J$ such that for $\{S_{n}\}\in I$,
$\prod_{n=1}^{V}P(S_{n}|H=0)\leq P_{F}'$ and for $\{S_{n}\}\in J$,
$\prod_{n=1}^{V}P(S_{n}|H=0)>P_{F}'$. The solution is given by the
rule that chooses $F=0$ for $\{S_{n}\}\in I$ and $F=1$ for $\{S_{n}\}\in J$.\\

For the Bayes fusion rule \cite{Ba95,Mi07}, the user specifies the
costs of a false alarm $C_{F}$, a missed detection $C_{M}$ and (for
$M>2$) a mix-up between two threat possibilities $C_{X}$. For each
combination of individual sensor decisions $\{S_{n}\}$, the system
finds a fused decision $F$ that minimizes the Bayes risk, or the
expected cost of a wrong decision,
\[
F=\mathrm{argmin}_{1\leq j\leq M}\sum_{k=1}^{M}C_{j,k}P(H=k-1)\prod_{n=1}^{V}P(S_{n}|H=k-1),
\]

where $C_{j,j}=0$, $C_{j,1}=C_{F}$ and $C_{1,j}=C_{M}$ for $j\geq2$,
and $C_{j,k}=C_{X}$ for all other $(j,k)$. The optimal fusion rule
can be found by simply looking at every combination $\{S_{n}\}$ individually
and taking the best of the $M$ possible fused decisions for each
one. In general, finding this fusion rule is a computationally difficult
discrete optimization problem, but this simple, brute-force approach
is fast as long as $V$ and $M$ are fairly small (e.g. less than
$10$), as is the case in our scenarios of practical interest.\\

\section{Dynamic fusion using multiple-stage sequential testing\label{SecDynamic}}

Suppose now that we have two static fusion networks of sensors, each
represented by a Bayesian network of the type described in Section
\ref{SecIntro}. The sensors collect measurements from a target moving
along a specified path, with the static network producing a fused
decision at every point in time based on the sensors' individual probabilities
at each position. These fused decisions can be combined over time
using Wald's theory of sequential probability testing. The classical
Wald sequential test is essentially a one-dimensional random walk
where the total likelihood ratio of the system makes ``steps'' in
either direction, corresponding to different incoming binary decisions.
The system reaches a fused decision when a specified upper or lower
threshold has been crossed. We refer to \cite{Po94} or \cite{Ve99}
for more details.\\

We develop an extension of the classical sequential test to cover
the following scenario. Only the sensors in the first static network
are initially active, and the sequential test accepts and combines
the fused outputs from that network at each point in time. If the
system detects an anomaly, it switches over to the second static network
and continues to pick up and combine measurements in the same manner
until a final decision has been reached. The motivation for this two-stage
setup is that there is typically a cost to activating and operating
the sensors in the second-stage network, so they are to be switched
on only if the first-stage sensors decide that there is a good chance
of a threat. The two networks do not need to be disjoint and can contain
some of the same sensors, although possibly with different graph linkages
or fusion rules. In practical scenarios, the second-stage graph is
a superset of the first-stage one that includes additional sensors,
reflecting the fact that the first-stage sensors continue to collect
observations after they trigger any additional sensors in the second-stage
network. It is also straightforward to add additional stages in the
same manner. For example, a third stage might correspond to an object
being acquired by video before sensors begin to collect measurements
on it (known as ``track before detect''). For clarity, however,
we focus on two stages in what follows.\\

Let $H$ be the object as before, and $D_{n}$ and $D_{n}'$ respectively
be the decision outputs of the first and second stage fusion networks
(at their respective final fusion centers) at time $n$. Assume that
the $\{D_{n}\}\bigcup\{D_{n}'\}$ are mutually independent. We restrict
$M=2$ for the rest of the paper, so the static fusion system gives
us the sequences of $2\times2$ matrices $P(D_{n}=\cdot|H)$ and $P'(D_{n}'=\cdot|H)$
for all times $1\leq n\leq N$. We use these inputs to set up the
following type of sequential test. We write $\underline{K}$, $\overline{K}$,
$K$, $\eta_{0}$ and $\eta_{1}$ for respectively the lower stopping
time, upper stopping time, stopping time, lower threshold and upper
threshold for the first stage of the test. The thresholds $\eta_{0}$
and $\eta_{1}$ are fixed parameters that control the overall sensitivity
of the test, while the stopping times are random variables that we
will specify below. Similarly, we write $\underline{K}'$, $\overline{K'}$,
$K'$, $\eta_{0}'$ and $\eta_{1}'$ for the corresponding variables
for the second stage of the test, where we require that $\eta_{0}'\leq\eta_{0}$
and $\eta_{1}'\geq\eta_{1}$. Let $\mathbf{D}_{k}=\{D_{n}\}_{1\leq n\leq k}$
denote the set of decisions of the system's active static fusion network
at each time $n$, up to time $k$. For any sequence of decision possibilities
$\mathbf{d}_{k}=\{d_{n}\}_{1\leq n\leq k}$, we define the likelihood
ratio recursively by
\[
L(\mathbf{d}_{k})=\left(\prod_{n=1}^{M(\mathbf{d}_{k})}\frac{P(D_{n}=d_{n}|H=1)}{P(D_{n}=d_{n}|H=0)}\right)\left(\prod_{n=M(\mathbf{d}_{k})+1}^{k}\frac{P'(D_{n}'=d_{n}|H=1)}{P'(D_{n}'=d_{n}|H=0)}\right),
\]

where $M(\mathbf{d}_{k})=\max\{n:n\leq k,L(\mathbf{d}_{m})\in(\eta_{0},\eta_{1})\,\forall\, m\in[1,n]\}$.
This allows us to define the stopping times by
\begin{eqnarray*}
\underline{K} & = & \min_{k\in[1,N]}\{k:L(\mathbf{D}_{k})\leq\eta_{0},L(\mathbf{D}_{m})\in(\eta_{0},\eta_{1})\,\forall\, m<k\},\\
\overline{K} & = & \min_{k\in[1,N]}\{k:L(\mathbf{D}_{k})\geq\eta_{1},L(\mathbf{D}_{m})\in(\eta_{0},\eta_{1})\,\forall\, m<k\},\\
K & = & \min(\underline{K},\overline{K}).\\
\underline{K}' & = & \min_{k\in[1,N]}\{k:k\geq K,L(\mathbf{D}_{k})\leq\eta_{0}',L(\mathbf{D}_{m})\in(\eta_{0}',\eta_{1}')\,\forall\, m<k\},\\
\overline{K}' & = & \min_{k\in[1,N]}\{k:k\geq K,L(\mathbf{D}_{k})\geq\eta_{1}',L(\mathbf{D}_{m})\in(\eta_{0}',\eta_{1}')\,\forall\, m<k\},\\
K' & = & \min(\underline{K}',\overline{K}').
\end{eqnarray*}

If any of the above sets is empty, we define the corresponding stopping
time to be $N+1$. The test starts with the first stage static network
$P(D_{n}=\cdot|H)$ and runs like a conventional sequential test until
either $k=K$ at time $k$ (i.e. when $L(\mathbf{D}_{k})$ moves outside
the region $(\eta_{0},\eta_{1})$), at which point it switches to
the second stage network, or until $k=N+1$, when it is forced to
stop and make a decision by comparing $L(\mathbf{D}_{N})$ to the
geometric midpoint $\sqrt{\eta_{0}\eta_{1}}$. If the second stage
is triggered, the test continues running with the second stage network
$P'(D_{n}'=\cdot|H)$ and stops with a final decision when $K'=k$
or $K'=N+1$. The first stage's result (representing an initial decision)
affects whether the second stage starts below $\eta_{0}$ or above
$\eta_{1}$. We want to find the statistics of $K$ and $K'$ and
the detection and false alarm probabilities of the first stage at
each time $k$, denoted $P_{D}^{k}$ and $P_{F}^{k}$, and of the
entire test, $P_{D}^{k}\,'$ and $P_{F}^{k}\,'$. \\

We describe a simple, deterministic algorithm to compute these quantities
for incoming, fused sensor measurements from the two static networks.
Let $G_{k}$ be the event $\{K\geq k\}$, i.e. that the first stage
was still running at time $k-1$. We can expand $P(\overline{K}=k|H,G_{k})$
for all $1\leq k\leq N$ by writing
\begin{eqnarray}
P(\overline{K}=k|H,G_{k}) & = & P(L(\mathbf{D}_{k})\geq\eta_{1}\,|\, L(\mathbf{D}_{m})\in(\eta_{0},\eta_{1}),1\leq m\leq k-1,H)\nonumber \\
 & = & \sum_{\mathbf{d}_{k}\in A_{k}(\eta_{0},\eta_{1})}\prod_{n=1}^{k}P(D_{n}=d_{n}|H),\label{ProbK}
\end{eqnarray}

where
\[
A_{k}(\eta_{0},\eta_{1})=\{\mathbf{d}_{k}\in\{0,1\}^{k}:L(\mathbf{d}_{k})\geq\eta_{1},L(\mathbf{d}_{m})\in(\eta_{0},\eta_{1}),1\leq m\leq k-1\}.
\]

We can express $P(\underline{K}=k|H)$ and the $k=N+1$ cases in a
similar manner. The number of terms in $A_{k}(\eta_{0},\eta_{1})$
generally grows exponentially, but the sum can be computed iteratively
by keeping track of likelihood sets $\mathcal{L}_{k}$ over time,
where each $\mathcal{L}_{k}$ consists of elements $\ell=(\ell_{1},\ell_{2})\in\mathbb{R}^{2}$.
For $k=1$, we set $\mathcal{L}_{1}:=\{P(D_{1}=d_{1}|H)\}$. For each
$k>1$, we find the likelihoods of all possible sample paths, $\mathcal{L}_{k}:=\{P(D_{k}=d_{k}|H)\times\ell:\ell\in\mathcal{L}_{k-1}\}$,
where $\times$ denotes the outer product between vectors in $\mathbb{R}^{2}$.
We then compute
\begin{eqnarray*}
P(\overline{K}=k|H,G_{k}) & = & \sum_{\ell\in\mathcal{L}_{k},\ell_{2}/\ell_{1}\geq\eta_{1}}\ell_{H+1}\\
P(\underline{K}=k|H,G_{k}) & = & \sum_{\ell\in\mathcal{L}_{k},\ell_{2}/\ell_{1}\leq\eta_{0}}\ell_{H+1},
\end{eqnarray*}

store the likelihoods of paths that escaped $\mathcal{M}_{k}:=\{\ell\in\mathcal{L}_{k}:\ell_{2}/\ell_{1}\not\in(\eta_{0},\eta_{1})\}$,
keep the remaining likelihoods $\mathcal{L}_{k+1}:=\mathcal{L}_{k}\backslash\mathcal{M}_{k}$
for the next step, and increment $k$. Note that at each time $k$,
we only need to keep $\mathcal{L}_{k}$ and $\mathcal{L}_{k-1}$ in
memory. At $k=N+1$, any likelihoods $\mathcal{L}_{N+1}$ still left
are summed over in a similar manner. From this, we can find
\begin{eqnarray*}
P(K=k|H) & = & P(\overline{K}=k|H,G_{k})+P(\underline{K}=k|H,G_{k}),\\
P_{D}^{k} & = & \sum_{m=1}^{k}P(\overline{K}=m|H=1,G_{k}),\\
P_{F}^{k} & = & \sum_{m=1}^{k}P(\overline{K}=m|H=0,G_{k}).
\end{eqnarray*}

In the same way, we can calculate the second stage probabilities.
For example, with $G_{k}'=\{K'<k\}$, we have
\begin{eqnarray*}
P(\overline{K'}=k|H,G_{k}') & = & \sum_{\mathbf{d}_{k}\in A_{k}(\eta_{1}')}\left(\prod_{n=1}^{M(\mathbf{d}_{k})}P(D_{n}=d_{n}|H)\right)\left(\prod_{n=M(\mathbf{d}_{k})+1}^{k}P'(D_{n}'=d_{n}|H)\right).
\end{eqnarray*}

These sums are evaluated in the same way as the first stage probabilities
by keeping sets of likelihoods $\mathcal{L}_{k}'$ in memory, with
the only differences being that for each $k\leq N$, we set $\mathcal{L}_{k}':=\mathcal{L}_{k}'\bigcup\mathcal{M}_{k}$
after computing $\mathcal{L}_{k}'$ as before (adding paths that cross
over between stages), and for $k=N+1$, we set $\mathcal{L}_{N+1}':=\mathcal{L}_{N+1}'\bigcup\mathcal{L}_{N+1}$.
Finally, we can calculate
\begin{eqnarray*}
E(K|H) & = & \sum_{k=1}^{N+1}kP(K=k|H),
\end{eqnarray*}

and any other statistics of $K$ and $K'$ can be found in the same
manner.\\

This iterative approach can provide a big performance improvement
over directly computing (\ref{ProbK}) in certain situations. The
likelihood set $A_{k}$ generally grows like $O(R^{k})$ for some
$R\in[1,2]$, but in practice, $R$ is fairly close to $1$ if either
$P(D_{n}=1|H=1)$ is increasing or $P(D_{n}=1|H=0)$ is decreasing,
which physically corresponds to the target moving closer to the sensor
network over time.\\

We finally remark that the thresholds $\eta_{0}$ and $\eta_{1}$
are selected in practice using the \textit{Wald approximations}, $\eta_{0}=\frac{1-P_{D}^{*}}{1-P_{F}^{*}}$
and $\eta_{1}=\frac{P_{D}^{*}}{P_{F}^{*}}$, for some target probabilities
$P_{D}^{*},P_{F}^{*}\in(0,1)$. It can be shown that at the mean stopping
time $k=E(K)$, $P_{D}^{k}\geq1-\frac{1-P_{D}^{*}}{1-P_{F}^{*}}$
and $P_{F}^{k}\leq\frac{P_{F}^{*}}{P_{D}^{*}}$ (\cite{Po94}, p.
104). The second stage thresholds $\eta_{0}'$ and $\eta_{1}'$ can
be chosen in a similar manner with some given $P_{D}^{*}\,'$ and
$P_{F}^{*}\,'$.

\section{Numerical examples\label{SecExamples}}

In this section, we consider a few example scenarios that illustrate
different properties of the static and dynamic elements described
above. To clarify the discussion, we consider a simple situation with
two sensors, $S_{1}$ and $S_{2}$, where the first-stage static network
includes only $S_{1}$, with no fusion, and the second-stage network
includes both $S_{1}$ and $S_{2}$ and combines them at a Bayes fusion
center with uniform costs and priors. At each time $k$ for $1\leq k\leq N$,
$N=25$, we assume the sensor $S_{j}$ has false alarm and detection
probabilities of $0.5-A_{j}-B_{j}k$ and $0.5+A_{j}+B_{j}k$ respectively,
where the $A_{j}$ and $B_{j}$ are some fixed constants and $j\in\{1,2\}$.
This setup loosely models a target object moving over time at a constant
speed, with only $S_{1}$ being initially active and triggering $S_{2}$
as needed. The thresholds are set using the Wald approximations as
described in Section \ref{SecDynamic}.\\

\begin{figure}[H]
\centering{}\includegraphics{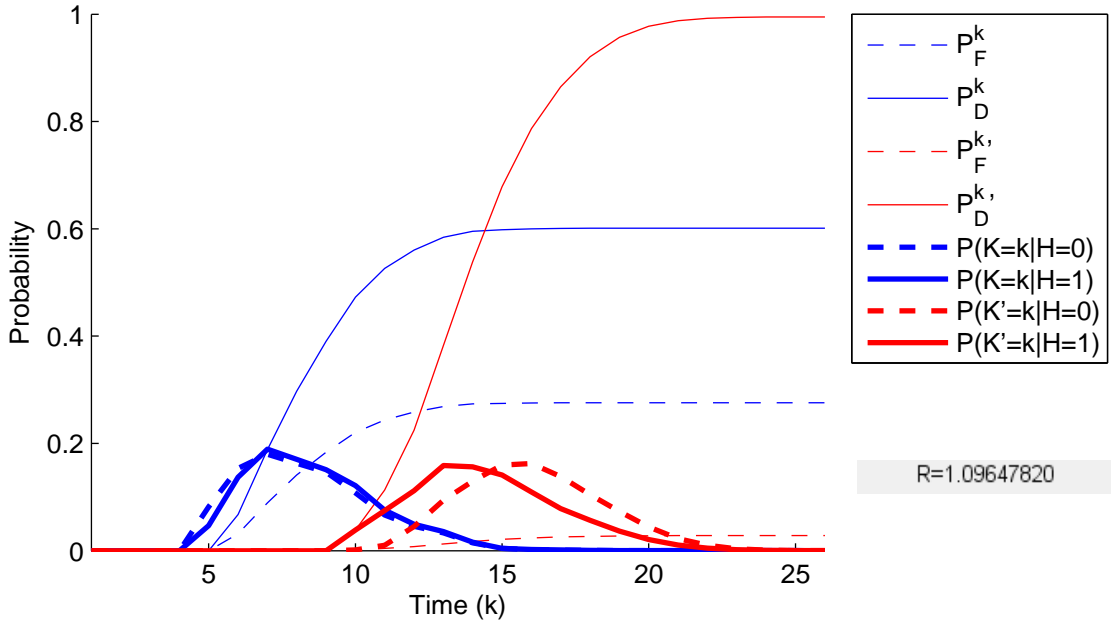}\caption{\label{FigSeq1}Fusion scenario with $\{A_{1},B_{1},A_{2},B_{2}\}=\{0,0.01,0,0.02\}$
and target probabilities $\{P_{F}^{*},P_{D}^{*},P_{F}^{*}\,',P_{D}^{*}\,'\}=\{0.3,0.55,0.05,0.99\}$.}
\end{figure}

\begin{figure}[H]
\centering{}\includegraphics{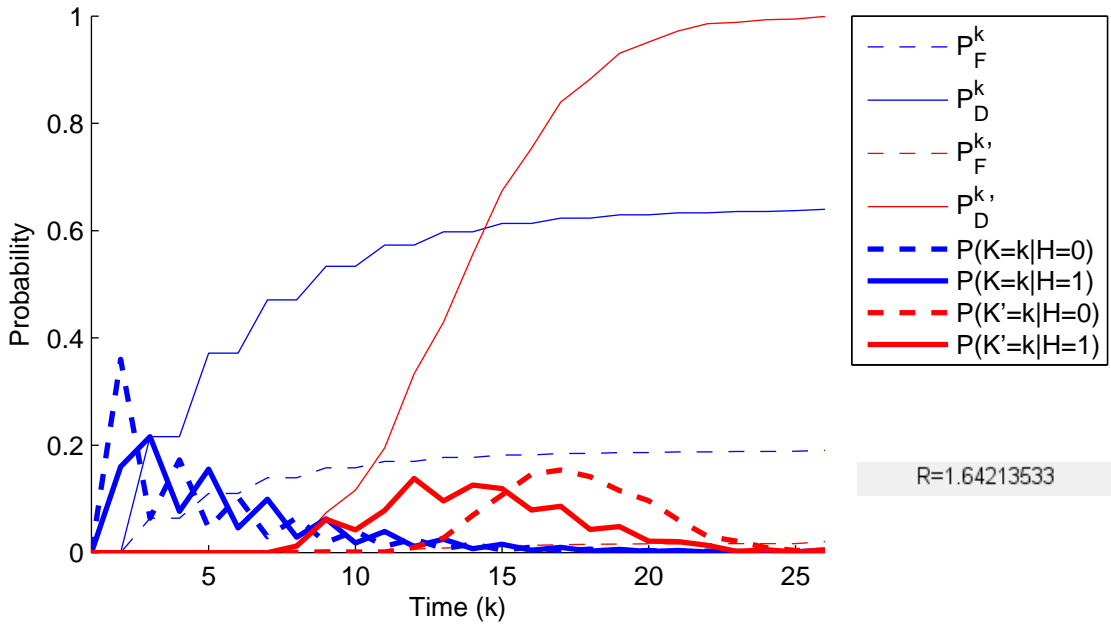}\caption{\label{FigSeq2}Fusion scenario with $\{A_{1},B_{1},A_{2},B_{2}\}=\{0.1,0,0,0.02\}$
and target probabilities $\{P_{F}^{*},P_{D}^{*},P_{F}^{*}\,',P_{D}^{*}\,'\}=\{0.2,0.55,0.03,0.999\}$.}
\end{figure}

The sequential test statistics are shown in Figures \ref{FigSeq1}
and \ref{FigSeq2} for various choices of the above parameters. The
first scenario corresponds to the target moving closer to both sensors
over time, while the second one describes a situation where the target
maintains a fixed distance from $S_{1}$, but moves towards $S_{2}$.
In both cases, the first-stage thresholds are set up to be modest
targets, so that $S_{1}$ quickly makes its preliminary decision and
activates $S_{2}$ well before the system reaches its final decision.
The stopping time distributions generally have large variances and
are highly oscillatory in the scenario with stationary sensor statistics,
but are smooth in the one with a moving target.\\

We also calculate estimates of the base $R$ in Section \ref{SecDynamic},
based on the number of sample paths still present at the final time
$N=25$. This gives an indication of how much computation time was
saved by culling out the sample paths that crossed the thresholds
at each time step. We find that $R$ is significantly less than $2$,
especially in the first case, and the running time is several orders
of magnitude less than it would be if we computed (\ref{ProbK}) directly.

\section{Conclusion}

We have described a general framework for decision-level data fusion
performance and tradeoff analysis between different sensor configurations.
We have discussed an extension of the classical Wald sequential test
to cover a multiple-stage cueing scenario where the decisions from
one sensor network are used to activate a second network for a closer
look at a target. We have described some numerical examples illustrating
the behavior of the resulting statistical quantities. These results
motivate future work on better characterizing the stopping time distributions
as well as the dependencies between the first and second stage times.

\bibliographystyle{plain}
\bibliography{Fullbib}

\begin{thebibliography}{1}

\bibitem{Ba95}
W.~Baek.
\newblock {Optimal m-ary data fusion with distributed sensors}.
\newblock {\em IEEE Trans. Aerospace and Electronic Systems}, 31(3), 1995.

\bibitem{Mi07}
H.~B. Mitchell.
\newblock {\em {Multi-sensor Data Fusion: An Introduction}}.
\newblock Springer, 2007.

\bibitem{Pe85}
J.~Pearl.
\newblock {Bayesian networks: a model of self-activated memory for evidential
  reasoning}.
\newblock {\em 7th Conference of the Cognitive Science Society}, 1985.

\bibitem{Pe97}
J.~Pearl.
\newblock {Bayesian Networks}.
\newblock {\em TR R-246, MIT Encyclopedia of the Cognitive Sciences}, 1997.

\bibitem{Po94}
H.~V. Poor.
\newblock {\em {An Introduction to Signal Detection and Estimation}}.
\newblock Springer, 1994.

\bibitem{Va96}
P.~K. Varshney.
\newblock {\em {Distributed Detection and Data Fusion}}.
\newblock Springer, 1996.

\bibitem{Ve99}
V.~V. Veeravalli.
\newblock {Sequential decision fusion: theory and applications}.
\newblock {\em Journal of the Franklin Institute}, 336:301--322, 1999.

\end{thebibliography}

\end{document}